\documentclass[lettersize,journal]{IEEEtran}
\usepackage{amsmath,amsfonts}
\usepackage{algorithmic}
\usepackage{algorithm}
\usepackage{array}
\usepackage[caption=false,font=normalsize,labelfont=sf,textfont=sf]{subfig}
\usepackage{textcomp}
\usepackage{stfloats}
\usepackage{url}
\usepackage{verbatim}
\usepackage{graphicx}
\usepackage{cite}
\usepackage{graphicx}
\usepackage{epstopdf}
\usepackage{multirow}
\usepackage{multicol}

\usepackage{booktabs}
\usepackage{bbding}

\usepackage[colorlinks,linkcolor=blue,citecolor=blue]{hyperref}

\begin{document}

\title{Improving Low-Latency Learning Performance in Spiking Neural Networks via a Change-Perceptive Dendrite-Soma-Axon Neuron}

\author{\IEEEauthorblockN{
		Zeyu Huang\IEEEauthorrefmark{1}, 
		Wei Meng\IEEEauthorrefmark{1}, 
		Quan Liu\IEEEauthorrefmark{1}, 
		Kun Chen\IEEEauthorrefmark{1} 
		and Li Ma\IEEEauthorrefmark{1}\IEEEauthorrefmark{2}} 
        \\
	\IEEEauthorblockA{\IEEEauthorrefmark{1}School of Information Engineering, Wuhan University of Technology, Wuhan, China}

\thanks{\IEEEauthorrefmark{2} Corresponding author. excellentmary@whut.edu.cn.}
\thanks{This work was supported by National Natural Science Foundation
of China (62572364), Natural Science Foundation of Wuhan Municipality (2024040801020263), Science and Technology Talents Serving Enterprise Program of Hubei Province (2025DJB030).
}}



\maketitle

\begin{abstract}
Spiking neurons, the fundamental information processing units of Spiking Neural Networks (SNNs), have the all-or-zero information output form that allows SNNs to be more energy-efficient compared to Artificial Neural Networks (ANNs). However, the hard reset mechanism employed in spiking neurons leads to information degradation due to its uniform handling of diverse membrane potentials. Furthermore, the utilization of overly simplified neuron models that disregard the intricate biological structures inherently impedes the network's capacity to accurately simulate the actual potential transmission process. To address these issues, we propose a dendrite-soma-axon (DSA) neuron employing the soft reset strategy, in conjunction with a potential change-based perception mechanism, culminating in the change-perceptive dendrite-soma-axon (CP-DSA) neuron. Our model contains multiple learnable parameters that expand the representation space of neurons. The change-perceptive (CP) mechanism enables our model to achieve competitive performance in short time steps utilizing the difference information of adjacent time steps. Rigorous theoretical analysis is provided to demonstrate the efficacy of the CP-DSA model and the functional characteristics of its internal parameters. Furthermore, extensive experiments conducted on various datasets substantiate the significant advantages of the CP-DSA model over state-of-the-art approaches.
\end{abstract}


\section{Introduction}
\IEEEPARstart{S}{piking} Neural Networks (SNNs) leverage asynchronous spikes for low energy information transmission\cite{maass1997networks}, attracting substantial research interest. Another ad-vantage of SNNs is the ability to process both spatial and temporal domain information, making them suitable for tasks such as video recognition\cite{miki2023spike}.

First, training an efficient and robust spiking neural network is a critical problem that has garnered the attention of many researchers. Currently, there are three main learning algorithms for SNNs training. One is Spike-Timing-Dependent Plasticity (STDP), which influences the strength of synapses through the temporal relationship of pre- and postsynaptic spikes. SNNs trained based on STDP still perform poorly due to the local optimization rule without global guided error compared with the back-propagation algorithm\cite{dong2023unsupervised}. Another is ANN-SNN conversion learning\cite{yan2021near}, which converts the pretrained ANN model to the SNN model. Conversion learning can achieve competitive results, but it has to consume a large number of time steps, which results in a significant amount of energy consumption. Moreover, conversion learning cannot utilize temporal domain information, making it impossible to train neuromorphic datasets. The other is direct supervised learning\cite{wu2018spatio}. The training algorithm employed in this study is Spatio-Temporal Backpropagation (STBP). This direct training approach exhibits robustness, enabling it to attain favorable training performance without any additional complicated skill. However, directly training deeper spiking neural networks faces the problem of gradient vanishing/exploding, as the spikes in the backpropagation process are non-differentiable, leading to inaccurate gradient propagation when using surrogate functions.
\begin{figure}[t]
\centering
\includegraphics[width=\columnwidth]{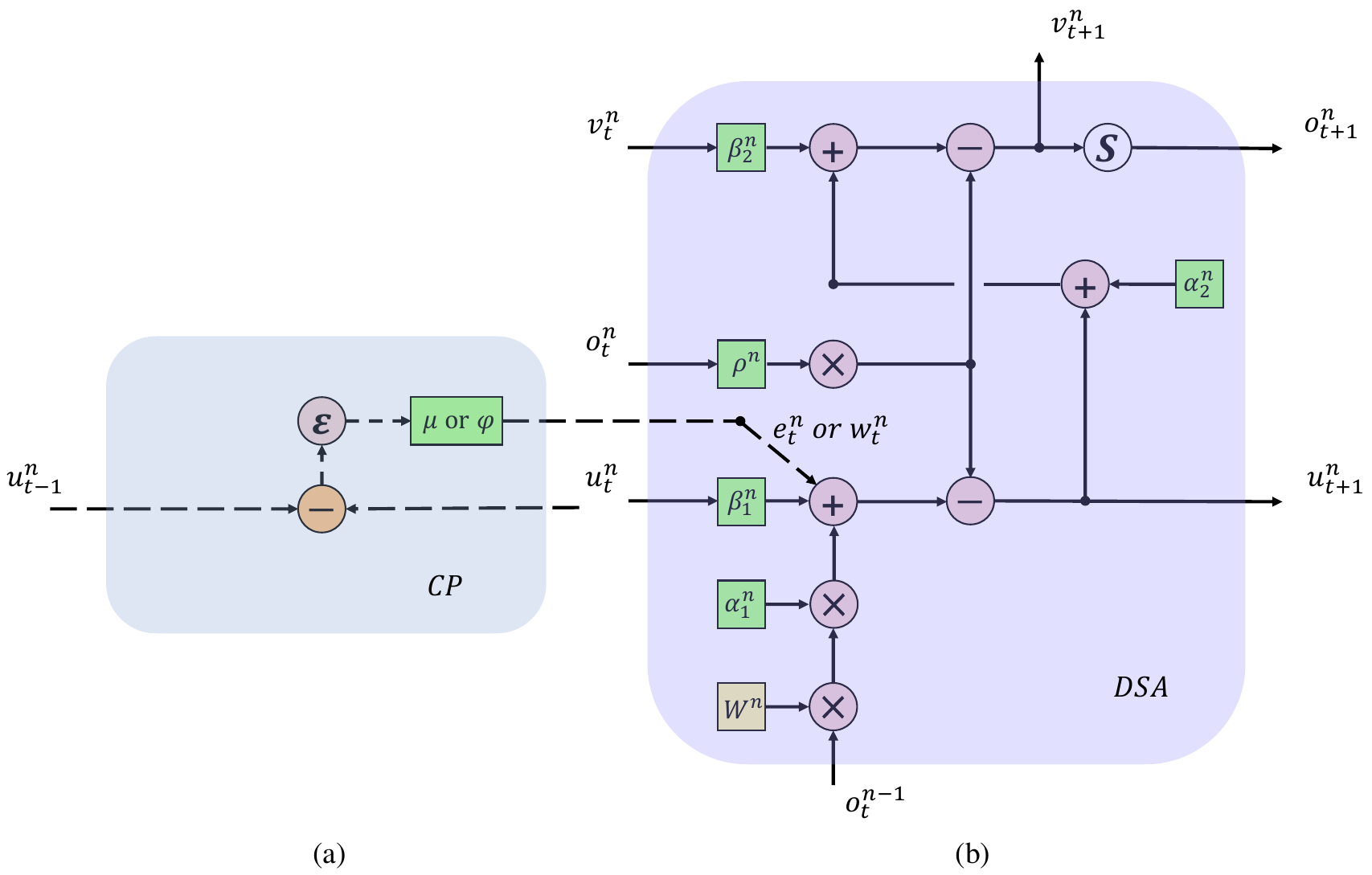} 
\caption{Illustration of the CP-DSA model. (a) Change-Perception mechanism. 
$\varepsilon$ denotes the step function. The subtraction operation determines the decrement and subtraction according to the CP mechanism. (b) Dendrite-Soma-Axon model. $S$ denotes output spike function.}
\label{fig1}
\end{figure}
The threshold-dependent batch normalization (tdBN) method in STBP-tdBN\cite{zheng2021going} addresses the gradient vanishing/exploding problem in direct training of deep spiking neural networks by satisfying the "Block Dynamical Isometry"\cite{chen2020comprehensive} and balancing threshold-input. Additionally, the multi-level firing (MLF) method\cite{feng2022multi} mitigates this issue by expanding the non-zero area of the rectangular approximate derivatives. Our model leverages the STBP-tdBN and MLF methods to mitigate the gradient vanishing/exploding issue.

Second, the neuron model is crucial to the performance of the entire SNN network, serving as the basic unit that processes and transmits input information. The Leaky Integrate-and-Fire (LIF) model is widely used in different SNNs\cite{wu2019direct,zheng2021going,fang2021deep,deng2022temporal,li2021differentiable}. The iterative LIF model proposed by the STBP method accelerates SNN training. To break through the performance bottleneck of neuron models, numerous variants of the LIF model have been proposed, such as the parametric LIF (PLIF) model\cite{fang2021incorporating}, the gated LIF (GLIF) model\cite{yao2022glif}, the adaptive self-feedback and balanced excitatory–inhibitory neuron\cite{zhao2022backeisnn}, etc. These variants are all single-compartment spiking neuron models that only retain the essential neuronal dynamics, ignoring the complex geometric structures of dendrites and axons and their interactions with the soma. Although single-compartment spiking neuron models greatly reduce the modeling effort, their performance on many complex tasks is not ideal, as the oversimplified single-compartment structure leads to information loss.

Inspired by the two-compartment LIF (TC-LIF) model\cite{zhang2024tc}, we re-evaluate the biological structure of neurons and provide a complete mechanistic equation to address the information loss caused by LIF. Unlike TC-LIF, our focus is not on the gradient vanishing problem induced by BPTT\cite{bird2021backpropagation} in the temporal dimension but on the biological structure itself. We propose spatial dimension modeling of the simplified biological neuron connection structure, namely the dendrite-soma-axon (DSA) structure. The DSA model contains multiple trainable parameters, as shown in Fig. \ref{fig1}(b). $u_t^n$, $v_t^n$, and $o_t^n$ represent the dendritic potential, somatic potential, and output spike of the n-th layer of the DSA model at time step $t$, respectively. $W^n$ denotes the weight of the n-th layer. $\alpha_1^n$ and $\alpha_2^n$ are the transfer coefficients from axon to dendrite and from dendrite to soma, respectively, which are used to control the intensity of the input. $\beta_1^n$ and $\beta_2^n$ are the decay coefficients of the potential on the dendrite and soma, respectively. All of these parameters are trainable. Each layer of the DSA is independent, which ensures the specificity of the training units in the network.

Most current research focuses on proposing modules to process input data, such as tdBN\cite{zheng2021going}, multi-dimensional attention SNN\cite{yao2023attention}, and the gated attention encoding (GAE) module\cite{qiu2024gated}. These modules extract information from data in various dimensions to facilitate neuron recognition. However, this paper focuses on the changes in input information that previous work has overlooked, quantifying these changes within the DSA model. We term this information change as the neuron's change-perceptive mechanism and use it to increase the proportion of activated neurons. As shown in Fig. \ref{fig1}(a), the step function was used to quantify the change of dendrites potential in adjacent time steps, multiplied by the learning factor, and the positive (or negative) increment $e_t^n$ ($w_t^n$) was introduced into the calculation of the DSA model. 

The main contributions of this paper are as follows:

\begin{itemize}
    \item Proposing a change-perceptive mechanism that quanti-fies changes in dendritic potential, affecting neuron fir-ing rates in the forward process and regulating spatio-temporal information propagation in the backward process.
    \item Introducing the CP-DSA model, which incorporates the change-perception mechanism. It is the simplest model of neuron structure at the potential level, offering multiple learnable parameters for adaptive adjustment during training, closely mirroring biological reality.
    \item Our experiments demonstrate the effectiveness of the CP-DSA model on three neuromorphic datasets. For ex-ample, our model can achieve the state-of-the-art 81.8\% top-1 accuracy on CIFAR10-DVS with only five time steps.
\end{itemize}

\section{Related Work}
\subsection{Diverse Neuron Models}
The basic connection structure for information transmission between neurons has inspired research on spiking neural networks. While the vanilla LIF model provides stable and convergent neuron dynamics, its overly simplified mechanism leads to poor performance on complex tasks. This has prompted researchers to explore LIF variants, establishing more intricate mechanisms for improved performance. PLIF model\cite{fang2021incorporating} increases individual neuron differences by making the decay coefficient a learnable parameter. GLIF\cite{yao2022glif} expands the representation space of neurons by incorporating biological features from different neuron behaviors. The multi-level firing (MLF) method\cite{feng2022multi} addresses gradient vanishing in deep networks by increasing spike firing rates through parallel neurons with different threshold levels. The ternary spike model\cite{guo2024ternary} introduces ternary spikes to compensate for information loss caused by binary spikes. TC-LIF, a biologically inspired two-compartment neuron model, addresses gradient vanishing in long-time step error propagation\cite{zhang2024tc}. The LM-H model\cite{hao2023progressive}, inspired by TC-LIF, dynamically optimizes membrane parameters to expand the computational range. Unlike the LIF variants that remain confined to the neuron itself, TC-LIF and LM-H expand the model's mechanism to encompass the dendrite-neuron structure. Our model is inspired by this concept, but it concentrates on the structural
time step, instead of across different time steps. We utilize change-perceptive mechanism to manage temporal information changes, quantifying it as part of our model.
\subsection{Training Algorithms for SNNs}
The STBP algorithm addresses the issue of non-differentiable spikes in backpropagation by employing surrogate gradient functions\cite{wu2018spatio}. The tdBN method suitable for SNN learning, making it possible to directly train deep SNNs\cite{zheng2021going}. Researchers noticed that the fixed width of surrogate gradients led to gradient vanishing and mismatch problems and proposed a learnable surrogate gradient method\cite{lian2023learnable} to effectively alleviate gradient propagation blockages when directly training deep SNNs. Some researchers focused on the definition of the loss function, they introduce the temporal efficient training (TET) method\cite{deng2022temporal} to compensate for the loss of momentum in the gradient descent with surrogate gradient, which helps the training process converge into flatter minima with better generalizability. Some have proposed attention-based modules, such as the GAE module\cite{qiu2024gated} and the advanced spatial attention (ASA) module\cite{yao2023inherent}. These modules essentially involve a series of ordered convolution operations, addition operations, and multiplication operations on the input raw data. These modules can help the network better extract spatiotemporal information, thereby improving the overall performance of the network. The other proposed a more refined spike-based ResNet structure\cite{fang2021deep} that supports three types of operations, enhancing the performance of spike network training. Our model uses the STBP method for training. The spike-based ResNet structure we use replaces the BN module with the tdBN module.

\begin{figure}[t]
\centering
\includegraphics[width=0.9\columnwidth]{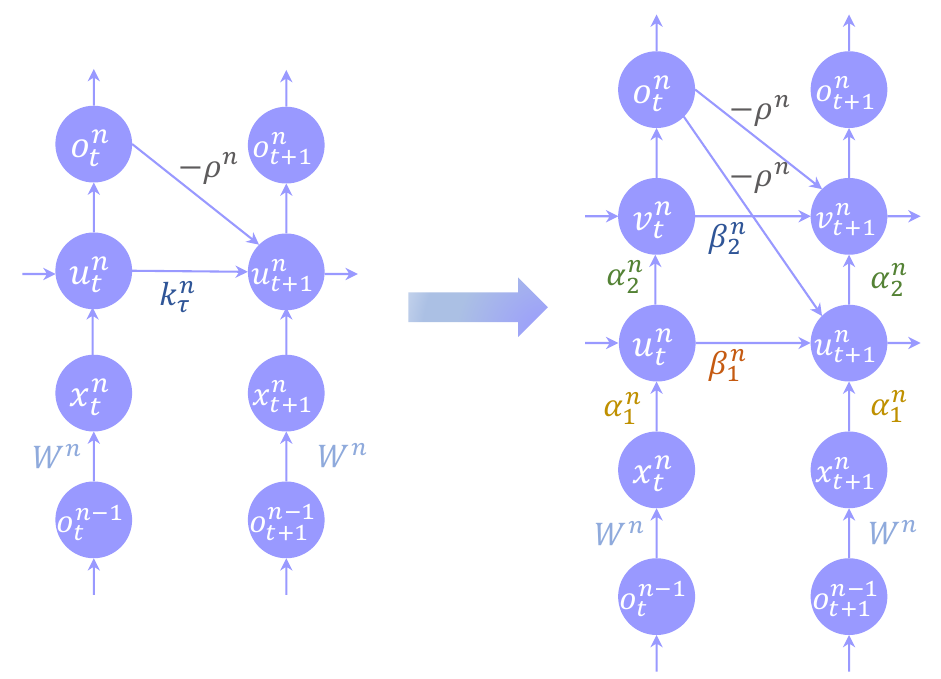} 
\caption{Evolution of the STBP training structure from soft-reset LIF neuron to DSA neuron.}
\label{fig2}
\end{figure}

\section{Method}
In this section, we first discuss the limitations of the LIF model. Then we propose the multi-parameter learnable dendrite-soma-axon (DSA) model inspired by the neuronal connection structure. Secondly, we propose a perception mechanism based on dendritic potential changes, and apply this mechanism to the DSA model, resulting in the CP-DSA model. By deriving the backpropagation process of CP-DSA, we theoretically analyze the role of each parameter and prove the effectiveness of our model in alleviating the gradient vanishing problem.
\subsection{Limitations of the LIF Model}
LIF is considered the most widely used and effective spiking neuron model, especially with the emergence of its iterative version\cite{fang2021deep}, which has accelerated its application in SNNs. However, analysis of experimental results indicates that the iterative LIF model's performance on certain neuromorphic tasks is suboptimal, which is related to its inherent dynamical mechanism. The discrete form of the iterative LIF neuron's dynamics can be described as:	
\begin{equation}
\label{eq1}
u_i^{t+1, n}= \begin{cases}\left(1-\frac{1}{\tau}\right) u_i^{t, n} \odot\left(1-o_i^{t, n}\right)+x_i^{t+1, n},&\text{hard reset} \\ \left(1-\frac{1}{\tau}\right) u_i^{t, n}-V_{\mathrm{th}} o_i^{t, n}+x_i^{t+1, n},&\text{soft reset}\end{cases}
\end{equation}
\begin{equation}
o_i^{t+1, n}=\Theta\left(u_i^{t+1, n}-V_{\text {th }}\right)
\end{equation}
\begin{equation}
x_i^{t+1, n}=\sum_j^{l(n-1)}\omega_{j i}^n o_j^{t+1, n-1}
\end{equation}
where $u_i^{t+1,n}$ and $x_i^{t+1,n}$ represent the membrane potential and input potential of the  i-th neuron in the  n-th layer at time  $t+1$, respectively. $\tau$  is the membrane time constant, and $\Theta$ represents the step function.

As shown in \eqref{eq1}, most of the current LIF models adopt a hard reset mechanism. When $o_i^{t,n}=1$, \eqref{eq1} becomes $u_i^{t+1,n}=x_i^{t+1,n}$. The potential that meets the threshold will reset to zero. The issue with hard reset mechanism is that no matter what the value of $u_i^{t+1,n}$ is, as long as it is greater than the threshold, it is considered the same information, causing the neuron to fire a spike and reset the potential to zero. This uniform mechanism to potential reset results in significant information loss.

On the other hand, the dynamics equation of the LIF model only describes the membrane potential update and spike generation mechanism, ignoring the structures involved in biological neuron interactions (such as axons, dendrites, and soma). Due to the simplification of dynamics, it becomes challenging for LIF neuron to fully process the acquired information.This limits the spatiotemporal processing power of the neuronal model.
\subsection{Multi-Parameter Learnable Dendrite-Soma-Axon (DSA) Model}
Inspired by neuronal connection structure, we attempt to introduce a more detailed working mechanism into the simplified neuron model (LIF). The connections between neurons are not simply potential transfers, the transmission process can be described as soma-axon-dendrite-soma. The ion concentration difference inside and outside the axon can compensate for the potential loss during axon transmission, but the transmission process from the axon terminal to the next neuron is not lossless and should not be ignored. Numerically, the potential transfer can be simplified to soma-axon terminal-dendrite-soma. The LIF model simplifies the potential transfer between the dendrite and soma, leading to inaccurate information transmission between neurons as it ignores potential loss during dendritic transmission. Based 
on this, we propose a dendrite-soma-axon model with a more complete transmission mechanism, and the discrete form of its dynamic equation can be described as:
\begin{equation}
\label{eq4}
u_i^{t+1, n}=\alpha_1 x_i^{t+1, n}+\beta_1 u_i^{t, n}-\rho o_i^{t, n}
\end{equation}
\begin{equation}
\label{eq5}
v_i^{t+1, n}=\alpha_2 u_i^{t+1, n}+\beta_2 v_i^{t, n}-\rho o_i^{t, n}
\end{equation}
\begin{equation}
o_i^{t, n}=\Theta\left(v_i^{t, n}-V_\text{th}\right)
\end{equation}
\begin{equation}
x_i^{t+1, n}=\sum_j^{l(n-1)} \omega_{j i}^n o_j^{t+1, n-1}
\end{equation}
where $u$ represents the potential on the dendrite, $v$ represents the potential on the soma, $x$ represents the input potential from soma of the front layer to dendrite of the current layer, $\alpha_1$ represents the transmission coefficient of the potential from axon terminal to dendrite. $\beta_1$ represents the decay coefficient of the dendritic potential, $\rho$ represents the soft reset coefficient, which controls the change of dendritic and somatic potentials after a spike is released, $\alpha_2$ represents the transmission coefficient of potential from dendritic to soma, $\beta_2$ represents the decay coefficient of the somatic potential, $\Theta$ represents the step function, $V_\text{th}$ represents the somatic potential threshold. 

When the somatic potential reaches the threshold for spike release, the dendritic and somatic potentials should change synchronously, meaning a soft reset operation is performed on both dendritic and somatic potentials simultaneously.

By substituting \eqref{eq4} into \eqref{eq5}, we obtain:

\begin{equation}
\label{eq8}
v_i^{t+1, n}=\underbrace{\beta_2 v_i^{t, n}}_{\text {Somatic}}+\underbrace{\alpha_2 \beta_1 u_i^{t, n}}_{\text {Dendritic }}+\underbrace{\alpha_1 \alpha_2 x_i^{t+1, n}}_{\text {Scaling }}-\underbrace{\alpha_2 \rho o_i^{t+1, n}}_{\text {Adaptive Reset }}
\end{equation}

As shown in the \eqref{eq8}, the somatic membrane potential at the current time step is composed of four components. The first term represents somatic memory, the second term represents dendritic memory, the third term represents the scaling of the input information at the current time step, and the fourth term corresponds to the adaptive soft reset. Compared to the soft reset membrane potential equation in \eqref{eq1}, DSA has more learnable parameters and greater expressive power. The introduction of dendritic potential extends the neuron’s temporal memory capacity.

Fig. \ref{fig2} illustrates the training structure evolving from LIF neuron to DSA neuron. Our model is trained using the STBP method, with all colored parameters in the figure set as trainable parameters.
\subsection{Change-Perceptive (CP) Mechanism}
Currently, researchers are focused on the neurons them-selves and the processing of input data, while neglecting the changes in neuron input information over time. We believe that the information received by neurons should not be limited to static information at the same time step, but should also consider the impact of changes in information over time on neurons. We describe this change-based perception mechanism as:
\begin{equation}
\label{eq9}
e_i^{t+1, n} =\mu \Theta\left(u_i^{t+1, n}-u_i^{t, n}\right)\odot u_i^{t+1, n}
\end{equation}
\begin{equation}
\label{eq10}
w_i^{t+1, n} =\varphi \Theta\left(u_i^{t, n}-u_i^{t+1, n}\right)\odot u_i^{t+1, n}
\end{equation}
where $\mu$ represents the scaling factor for the enhancement mechanism,  $\varphi$ represents the scaling factor for the weaken mechanism, and $\Theta$ represents the step function.

\eqref{eq9} describes the increment brought by positive changes in dendritic potential, while \eqref{eq10} describes the increment brought by negative changes in dendritic potential. We believe that stronger stimuli make neurons more likely to fire spikes. By quantifying the positive increment of dendritic potential as described in \eqref{eq9}, we can implement the change-perception mechanism by influencing the dendritic potential change. Similarly, weakened stimuli should not be ignored, and we quantify them as described in \eqref{eq10}, thereby affecting the dendritic potential.

\begin{figure}[t]
\centering
\includegraphics[width=0.9\columnwidth]{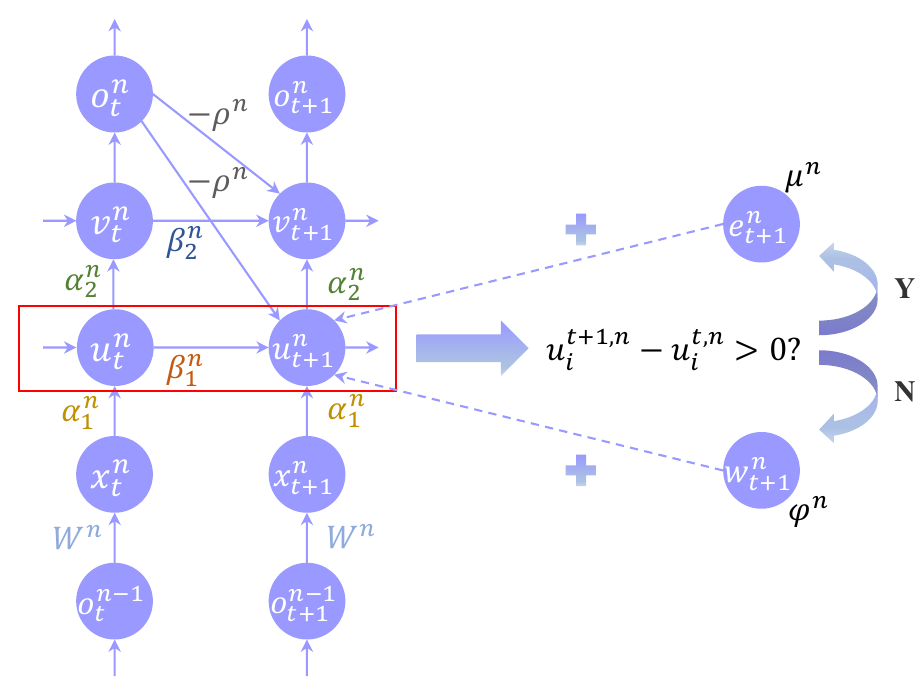} 
\caption{The impact of the change perceptive mechanism on the DSA training process.}
\label{fig3}
\end{figure}

The impact of the change-perceptive (CP) mechanism on the training process of DSA neurons is illustrated in Fig. \ref{fig3}. CP determines whether the information is enhanced or weakened by evaluating the differences in dendritic potential between adjacent time steps. According to \eqref{eq9} and \eqref{eq10}, it calculates the result and linearly adds it to the dendritic potential to influence the training process of DSA.

\subsection{Multi-Parameter Learnable Dendrite-Soma-Axon Neuron with the Change-Perceptive Capability (CP-DSA) Model}
The dendrite-soma-axon model enhances the neuron model's expressiveness by incorporating the dynamic equations of dendritic potential and somatic potential. The dendrite-soma-axon model can be seen as a simplified yet complete neuron structure in terms of potential hierarchy. We introduce the change-perceptive mechanism into the dendrite-soma-axon model, and its forward propagation can be described as:
\begin{equation}
\label{eq11}
\left\{\begin{array}{c}
u_i^{t+1, n}=\alpha_1 x_i^{t+1, n}+\beta_1 u_i^{t, n}-\rho o_i^{t, n} \\
v_i^{t+1, n}=\alpha_2 u_i^{t+1, n}+\beta_2 v_i^{t, n}-\rho o_i^{t, n}+e_i^{t+1, n} \\
o_i^{t, n}=\Theta\left(v_i^{t, n}-V_\text{th}\right)
\end{array}\right.
\end{equation}
\begin{equation}
\label{eq11}
\left\{\begin{array}{c}
\tilde{u}_i^{t+1, n}=\alpha_1 x_i^{t+1, n}+\beta_1 \tilde{u}_i^{t, n}-\rho \tilde{o}_i^{t, n} \\
\tilde{v}_i^{t+1, n}=\alpha_2 \tilde{u}_i^{t+1, n}+\beta_2 \tilde{v}_i^{t, n}-\rho \tilde{o}_i^{t, n}+w_i^{t+1, n} \\
\tilde{o}_i^{t, n}=\Theta\left(\tilde{v}_i^{t, n}-U_\text{th}\right) \\
\end{array}\right.
\end{equation}
\begin{equation}
\label{eq12}
s_i^{t+1, n}=o_i^{t+1, n}+\tilde{o}_i^{t+1, n}
\end{equation}
\begin{equation}
x_i^{t+1, n}=\sum_j^{l(n-1)} w_{j i}^n s_j^{t+1, n-1}
\end{equation}
where \eqref{eq11} refers to neurons with enhanced perceptual potential, \eqref{eq12} refers to neurons with weakened perceptual potential, $s_i^{t+1,n}$  represents the  i-th neuron’s total spike in the  n-th layer at time step  $t+1$, $e_i^{t+1,n}$ and $w_i^{t+1,n}$ can be found in \eqref{eq9} and \eqref{eq10}.

The entire model consists of two dendrite-soma-axon (DSA) structures. The first DSA neuron focuses on enhanced information, while the second DSA neuron focuses on weaken information. Together, they form the output of the overall model. Two types of neurons have different thresholds.
\begin{table*}[t]
\centering
\caption{Training hyperparameters for CP-DSA.}
\begin{tabular}{c c c c}
\toprule
Dataset & CIFAR10-DVS & N-Caltech101 & DVS-Gesture \\
\midrule
Optimizer & SGD & SGD & SGD\\
Init learning rate & 0.1 & 0.1 & 0.1\\
Weight decay & 1e-3 & 1e-4 & 1e-4\\
Lr schedule & StepLR(0.1/40) & MultiStepLR(0.1/ [60,80]) & StepLR(0.1/80)\\
dropout & 0.2 & 0.2 & 0\\
Total epoch & 130 & 90 & 200\\
\bottomrule
\end{tabular}
\label{tab1}
\end{table*}
To demonstrate that our model can facilitate more effective gradient propagation, we derive the backpropagation process as follows:

Consider the loss function $L$. By using the chain rule, $\frac{\partial L}{\partial s_i^{t, n}}, \frac{\partial L}{\partial o_i^{t, n}}, \frac{\partial L}{\partial u_i^{t, n}}$ and $\frac{\partial L}{\partial v_i^{t, n}}$ can be calculated as follows: 
\begin{equation}
\begin{aligned}
\left.\frac{\partial L}{\partial s_i^{t, n}} \right\rvert\, & =\sum_{j=1}^{l(n+1)} \frac{\partial L}{\partial x_j^{t, n+1}} \frac{\partial x_j^{t, n+1}}{\partial s_i^{t, n}} \\
& =\sum_{j=1}^{l(n+1)} \frac{\partial L}{\partial x_j^{t, n+1}} \omega_{i j}^{n+1} \\
& =\sum_{j=1}^{l(n+1)} \omega_{i j}^{n+1}\big(\frac{\partial L}{\partial u_j^{t, n+1}} \frac{\partial u_j^{t, n+1}}{\partial x_j^{t, n+1}}+\frac{\partial L}{\partial \tilde{u}_j^{t, n+1}} \frac{\partial \tilde{u}_j^{t, n+1}}{\partial x_j^{t, n+1}}\big) \\
& =\sum_{j=1}^{l(n+1)} \omega_{i j}^{n+1} \alpha_1\left(\frac{\partial L}{\partial u_j^{t, n+1}}+\frac{\partial L}{\partial \tilde{u}_j^{t, n+1}}\right)
\end{aligned}
\end{equation}
\begin{equation}
\begin{aligned}
\frac{\partial L}{\partial o_i^{t, n}} & =\frac{\partial L}{\partial u_i^{t, n}} \frac{\partial u_i^{t, n}}{\partial o_i^{t, n}}+\frac{\partial L}{\partial v_i^{t, n}}+\frac{\partial v_i^{t, n}}{\partial o_i^{t, n}}+\frac{\partial L}{\partial s_i^{t, n}} \frac{\partial s_i^{t, n}}{\partial o_i^{t, n}} \\
& =\frac{\partial L}{\partial u_i^{t, n}}(-\rho)+\frac{\partial L}{\partial v_i^{t, n}}(-\rho)+\frac{\partial L}{\partial s_i^{t, n}}
\end{aligned}
\end{equation}
\begin{equation}
\label{eq17}
\begin{aligned}
\frac{\partial L}{\partial u_i^{t, n}}& =\frac{\partial L}{\partial u_i^{t+1, n}} \frac{\partial u_i^{t+1, n}}{\partial u_i^{t, n}}+\frac{\partial L}{\partial v_i^{t, n}} \frac{\partial v_i^{t, n}}{\partial u_i^{t, n}} \\
& =\frac{\partial L}{\partial u_i^{t+1, n}} \beta_1+\frac{\partial L}{\partial v_i^{t, n}}\left(\alpha_2+\mu \Theta\left(u_i^{t, n}-u_i^{t-1, n}\right)\right. \\
& \big.+\mu \frac{\partial \Theta\left(u_i^{t, n}-u_i^{t-1, n}\right)}{\partial u_i^{t, n}} u_i^{t, n}\big)
\end{aligned}
\end{equation}
\begin{equation}
\label{eq18}
\begin{aligned}
\frac{\partial L}{\partial v_i^{t, n}}& =\frac{\partial L}{\partial v_i^{t+1, n}} \frac{\partial v_i^{t+1, n}}{\partial v_i^{t, n}}+\frac{\partial L}{\partial o_i^{t, n}} \frac{\partial o_i^{t, n}}{\partial v_i^{t, n}} \\
& =\frac{\partial L}{\partial v_i^{t+1, n}} \beta_2+\frac{\partial L}{\partial o_i^{t, n}} \frac{\partial o_i^{t, n}}{\partial v_i^{t, n}}
\end{aligned}
\end{equation}

Use $f(x)\equiv0$ instead of $\frac{\partial \Theta\left(u_i^{t, n}-u_i^{t-1, n}\right)}{\partial u_i^{t, n}}$ to avoid the possibility of gradient explosion in \eqref{eq17}.

When $u_i^{t, n}-u_i^{t-1, n}>0$, \eqref{eq18} simplifies to:
\begin{equation}
\label{eq19}
\frac{\partial L}{\partial u_i^{t, n}}=\frac{\partial L}{\partial u_i^{t+1, n}} \beta_1+\frac{\partial L}{\partial v_i^{t, n}}\left(\alpha_2+\mu\right)
\end{equation}

When $u_i^{t, n}-u_i^{t-1, n}<0$, \eqref{eq17} simplifies to:
\begin{equation}
\label{eq20}
\frac{\partial L}{\partial u_i^{t, n}}=\frac{\partial L}{\partial u_i^{t+1, n}} \beta_1+\frac{\partial L}{\partial v_i^{t, n}} \alpha_2
\end{equation}

Similarly, the partial derivatives of the loss function $L$ with respect to the neurons processing attenuated information can be obtained in a similar manner.
\begin{table}[t]
\centering
\caption{Settings for CP-DSA parameters.}
\begin{tabular}{c c}
\toprule
Parameter & Value\\
\midrule
$\alpha_1$ & $1.0(0\sim2)$\\
$\alpha_2$ & $0.5(0\sim1)$\\
$\beta_1$ & $0.5(0\sim1)$\\
$\beta_2$ & $0.5(0\sim1)$\\
$\rho$ & $0.5(0\sim1)$\\
$\mu$ & $0.5(0\sim1)$\\
$\phi$ & $0.5(0\sim1)$\\
$V_\text{th}$ & $0.5$\\
$U_\text{th}$ & $1.0$\\
\bottomrule
\end{tabular}
\label{tab2}
\end{table}
From \eqref{eq19} and \eqref{eq20}, it is easy to see that the backpropagation of the dendritic potential $u$ consists of both temporal and spatial gradients. The impact of the change-perception mechanism on backpropagation is achieved by learnable factors that regulate the proportion of these temporal and spatial gradients.

Compared to the LIF model, the temporal and spatial gradients in our model can be adjusted by multiple parameters $\alpha$ and $\beta$. During the initial phases of training, they maintain a larger fluctuation, allowing the network to adjust parameters with significant changes. As training progresses, these parameters gradually stabilize, enabling the network to converge by making smaller adjustments, which allows our model to converge to a better result in a relatively short amount of time.

Due to the non-differentiability of spiking activity, the derivative of $\frac{\partial o_i^{t, n}}{\partial v_i^{t, n}}$ in \eqref{eq18} cannot be directly obtained. To address this issue, we use the rectangular function\cite{wu2018spatio} to approximate the derivative of the spiking activity, which is defined as follows:
\begin{equation}
\frac{\partial o^t}{\partial v^t} \approx \frac{1}{a} \operatorname{sign}\left(\left|u^t-V_\text{th}\right|<\frac{a}{2}\right)
\end{equation}
where $a$ is the width of the rectangular function.
Based on the calculation of the above partial derivatives, the partial derivatives of the loss function  $L$ with respect to the network weights can be computed as follows:
\begin{equation}
\label{eq22}
\begin{aligned}
\frac{\partial L}{\partial w_{j i}^n}& =\sum_{t=1}^T \frac{\partial L}{\partial x_i^{t, n}} \frac{\partial x_i^{t, n}}{\partial w_{j i}^n} \\
& =\sum_{t=1}^T \frac{\partial L}{\partial x_i^{t, n}} s_j^{t, n-1} \\
& =\sum_{t=1}^T \alpha_1\left(\frac{\partial L}{\partial u_i^{t, n}}+\frac{\partial L}{\partial \tilde{u}_i^{t, n}}\right)\left(o_i^{t+1, n}+\tilde{o}_i^{t+1, n}\right)
\end{aligned}
\end{equation}

Observing \eqref{eq22}, it can be found that the update of the weights by the loss function $L$ is influenced by the changes in dendritic potential and spike firing of both types of neurons. In comparison to the single neuron firing in the LIF model, it is easier to conclude that the double neuron spike firing makes it difficult for the gradient of $L$ with respect to the weights to be zero, which to some extent avoids gradient vanishing. Additionally, even with short time step $T$, the partial derivative values do not approach zero, which facilitates the rapid convergence of our model within a short time step. The introduction of $\alpha_1$, as a factor regulating the magnitude of the partial derivatives, prevents gradient explosion caused by the excessive accumulation of spikes.

\begin{table*}[t]
\centering
\caption{Structures of ResNet, where fc denotes the fully connected layer.}
\begin{tabular}{c c c c c c}
\toprule
Dataset & \multicolumn{2}{c}{CIFAR10-DVS} & DVS-Gesture & \multicolumn{2}{c}{N-Caltech101} \\
\midrule
Layer & ResNet-14 & ResNet-20 & ResNet-14 & ResNet-14 & ResNet-20\\
\midrule
$Conv1$ & \multicolumn{2}{c}{$3\times3,32$} & $3\times3,16$ & \multicolumn{2}{c}{$3\times3,64$}\\
\midrule
$Conv2\_x$ & $\begin{bmatrix}
3\times3,32 \\
3\times3,32
\end{bmatrix}\times2$ & $\begin{bmatrix}
3\times3,32 \\
3\times3,32
\end{bmatrix}\times3$ & $\begin{bmatrix}
3\times3,16 \\
3\times3,16
\end{bmatrix}\times2$ & $\begin{bmatrix}
3\times3,64 \\
3\times3,64
\end{bmatrix}\times2$ & $\begin{bmatrix}
3\times3,64 \\
3\times3,64
\end{bmatrix}\times3$ \\
\midrule
$Conv3\_x$ & $\begin{bmatrix}
3\times3,64 \\
3\times3,64
\end{bmatrix}\times2$ & $\begin{bmatrix}
3\times3,64 \\
3\times3,64
\end{bmatrix}\times3$ & $\begin{bmatrix}
3\times3,32 \\
3\times3,32
\end{bmatrix}\times2$ & $\begin{bmatrix}
3\times3,128 \\
3\times3,128
\end{bmatrix}\times2$ & $\begin{bmatrix}
3\times3,128 \\
3\times3,128
\end{bmatrix}\times3$ \\
\midrule
$Conv4\_x$ & $\begin{bmatrix}
3\times3,128 \\
3\times3,128
\end{bmatrix}\times2$ & $\begin{bmatrix}
3\times3,128 \\
3\times3,128
\end{bmatrix}\times3$ & $\begin{bmatrix}
3\times3,64 \\
3\times3,64
\end{bmatrix}\times2$ & $\begin{bmatrix}
3\times3,256 \\
3\times3,256
\end{bmatrix}\times2$ & $\begin{bmatrix}
3\times3,256 \\
3\times3,256
\end{bmatrix}\times3$ \\
\midrule
& \multicolumn{2}{c}{Average pool, 10-d, fc} & Average pool, 11-d, fc & \multicolumn{2}{c}{Average pool, 101-d, fc}\\
\bottomrule
\end{tabular}
\label{tab3}
\end{table*}

\section{Experiment}
\subsection{Dataset Introduction}
To validate the effectiveness of our proposed model and mechanism, we conduct experiments on three neuromorphic datasets, including CIFAR10-DVS\cite{li2017cifar10}, N-Caltech101\cite{orchard2015converting}, and DVS-Gesture\cite{amir2017low}.The preprocessing of neuromorphic data in our work is analogous to the methodology described in MLF\cite{feng2022multi}, which involves integrating the event streams into frames and subsequently downsampling. However, in contrast to their approach, our work leverages the spikingjelly\cite{fang2023spikingjelly} toolkit , rather than relying on custom processing functions. 

\textbf{CIFAR10-DVS:} CIFAR10-DVS is a neuromorphic version of the CIFAR10 dataset and serves as a benchmark for evaluating the performance of Spiking Neural Networks (SNNs). CIFAR10-DVS contains 10,000 samples, with each sample represented as an event stream. The spatial size of all samples is 128 × 128, and there are 10 classes of objects. Each event stream $s \in [t, x, y, p]$ indicates the change in pixel value or brightness at location $[x, y]$ at the moment t relative to the previous moment. The prepresents polarity, where positive polarity indicates an increase in pixel value or brightness, and vice versa.

The original event stream in the CIFAR10-DVS dataset is split into multiple slices in 100ms increments. Each slice is then integrated into a frame and downsampled to a spatial size of 48 × 48. The CIFAR10-DVS dataset is randomly divided into training and test sets, with a ratio of 9:1.

\textbf{N-Caltech101:} N-Caltech101 is a neuromorphic version of Caltech101 with 101 categories and 8709 samples with a spatial size of 180 × 240.

We use the entire length of the event stream, dividing it evenly based on the time step into T slices, which are then integrated into T frames and downsampled to a spatial size of 64x64. The N-Caltech101 dataset is randomly divided into training and test sets, with a ratio of 9:1.

\textbf{DVS-Gesture:} DVS-Gesture is a neuromorphic dataset designed for gesture recognition. It comprises 11 event stream samples of gestures, with 1176 samples for training and 288 for testing. Each sample has a spatial size of 128 × 128. In the DVS-Gesture dataset, the event stream is integrated into frames at intervals of 100ms and then downsampled to 48 × 48.

\subsection{Experimental Settings}
We choose the currently mainstream network architecture ResNet, which is consistent with the network architecture selected in the baseline method MLF\cite{feng2022multi}.The details of the network architecture can be found in Table\ref{tab1}. The initialization parameters of CP-DSA are shown in Table\ref{tab2}, where the threshold is a fixed value and the remaining parameters are adaptively adjusted within a specified range. Table\ref{tab3} presents the training hyperparameters, including the optimizer used for each dataset, the learning rate adjustment strategy, and other relevant information. The ablation study setup is consistent with the formal experiments, and the experiments are repeated three times to average the results and eliminate the effects of randomness.

\begin{figure*}[t]
\centering
\includegraphics[width=0.9\textwidth]{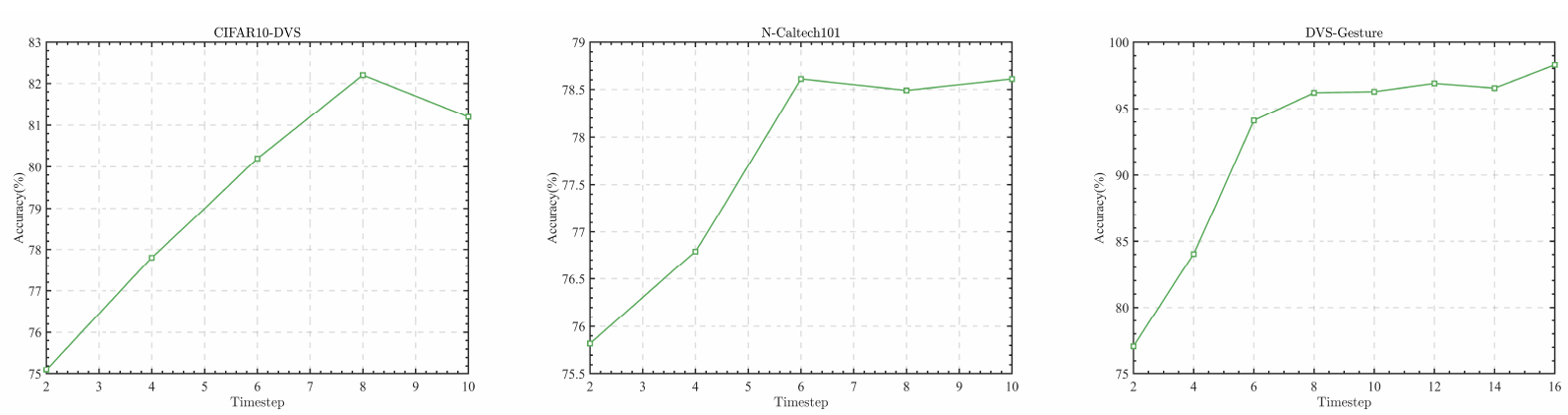} 
\caption{The performance of the CP-DSA model at different time steps.The circles correspond to experimentally measured data, which demonstrate that, even at very low time steps, CP-DSA still outperforms other methods listed in Table \ref{tab4}.}
\label{fig4}
\end{figure*}

\begin{table*}[t]
\centering
\caption{Comparison with previous SOTA works.}
\begin{tabular}{c c c c c}
\toprule
Dataset & Method & Architecture	& Timestep & Accuracy(\%) \\
\midrule
\multirow{6}{*}{CIFAR10-DVS} & STBP-tdBN\cite{zheng2021going} & ResNet-19 & 10 & 67.80\\
& MLF\cite{feng2022multi} & ResNet-20 & 10 & 70.36\\
& SSNN\cite{ding2024shrinking} & VGG-9 & 8 & 78.57\\
& LM-H\cite{hao2023progressive}	& ResNet-19	& 10 & 79.10\\
& Spike-driven Transformer\cite{yao2024spike} & Spiking Transformer & 16 & 80.0\\
& Ours & ResNet-14 & 5/8/10 & 81.8/\textbf{82.2}/81.2\\
\midrule
\multirow{7}{*}{N-Caltech101} & SALT\cite{kim2021optimizing} & VGG-11 & 20 & 55.0\\
& EventMix\cite{shen2023eventmix} & ResNet-18 & 10 & 79.47\\
& SSNN\cite{ding2024shrinking} & VGG-9 & 8 & 79.25\\
& TIM\cite{shen2024tim} & Spikformer & 10 & 79\\
& NDA\cite{li2022neuromorphic} & ResNet-19 & 10 & 78.6\\
& PFA\cite{deng2024tensor} & VGGSNN & 14 & 80.5\\
& Ours & ResNet-20 & 8/10 & 82.5/\textbf{83.47}\\
\midrule
\multirow{5}{*}{DVS-Gesture} & STBP-tdBN\cite{zheng2021going} & ResNet-19 & 40 & 96.9\\
& MLF\cite{feng2022multi} & ResNet-20 & 40 & 97.29\\
& SSNN\cite{ding2024shrinking} & VGG-9 & 8 & 94.91\\
& Spike-driven Transformer\cite{yao2024spike} & Spiking Transformer & 16 & \textbf{99.3}\\
& Ours & ResNet-14 & 8/16 & 96.18/98.26\\
\bottomrule
\end{tabular}
\label{tab4}
\end{table*}

\subsection{Comparison with Existing Works}
\subsubsection{CIFAR10-DVS}As we can see from Table \ref{tab4}, compared to previous methods, Spike-driven Transformer\cite{yao2024spike} model achieves the best performance on the CIFAR10-DVS dataset, reaching an accuracy of 80.0\%. However, it requires relatively longer time steps. In contrast, the SSNN\cite{ding2024shrinking} method obtains an excellent accuracy of 78.53\% using only 8 time steps. We apply ResNet-14 to evaluate the performance of our model. In order to provide a comparative analysis, we conduct experiments at 5, 8, and 10 time steps, and obtain state-of-the-art accuracies of 81.8\%, 82.2\%, and 81.2\%, respectively. Notably, at just 8 time steps, our method surpasses the Spike-driven Transformer\cite{yao2024spike} model by 2.2\%, and outperformed the SSNN\cite{ding2024shrinking} method by 3.63\% at the same time steps.
\subsubsection{N-Caltech101} We apply ResNet-20 to evaluate the performance of our model. As indicated in Table\ref{tab4}, our model achieves the optimal performance of 83.47\% accuracy at 10 time steps. This surpasses the PFA\cite{deng2024tensor} method by 2.97\%, and outperforms the SSNN\cite{ding2024shrinking} method by 3.25\% at 8 time steps.
\subsubsection{DVS-Gesture} Spike-driven Transformer\cite{yao2024spike} achieves the best performance at 16 time steps, with an accuracy of 99.3\%. Our model achieves an accuracy of 98.26\% at 16 time steps. Although slightly lower than the Spike-driven Transformer model, it outperformed the tdBN\cite{zheng2021going}, MLF\cite{feng2022multi}, and SSNN\cite{ding2024shrinking} models at shorter time steps.

\begin{table}[t]
\centering
\caption{Comparison with the baseline. The data in the table are self-implementation results.}
\begin{tabular}{c c c c c}
\toprule
Dataset & Method & Architecture	& Timestep & Accuracy(\%) \\
\midrule
\multirow{4}{*}{CIFAR10-DVS} & \multirow{2}{*}{MLF\cite{feng2022multi}} & ResNet-14 & 4 & 74.9\\
& & ResNet-20 & 4 & 74.7\\
& \multirow{2}{*}{Ours} & ResNet-14 & 4 & 77.7\\
& & ResNet-20	& 4 & \textbf{79.4}\\
\midrule
\multirow{2}{*}{N-Caltech101} & MLF\cite{feng2022multi} & ResNet-20 & 5 & 79.46\\
& Ours & ResNet-20 & 5 & \textbf{81.53}\\
\midrule
\multirow{2}{*}{DVS-Gesture} & MLF\cite{feng2022multi} & ResNet-14 & 10 & 95.83\\
& Ours & ResNet-14 & 10 & \textbf{96.83}\\
\bottomrule
\end{tabular}
\label{tab5}
\end{table}
\subsection{Comparison with the Baseline}
Since our model uses a multi-threshold level neuron structure to mitigate the vanishing gradient phenomenon, similar to MLF\cite{feng2022multi}, we choose MLF as the baseline for comparison. We use consistent data processing methods and conduct comparative experiments with the same network architecture and time steps. The results are shown in Table\ref{tab5}. On the CIFAR10-DVS dataset, at 4 time steps, the CP-DSA model outperforms the MLF method in accuracy, whether using ResNet-14 or ResNet-20. On the ResNet-20, it exceeds by 4.7\%. On the N-Caltech101 dataset, the CP-DSA model's accuracy surpasses the MLF method by 2.07\%. On the DVS-Gesture dataset, the CP-DSA model's accuracy exceeds the MLF method by 1\%.

\subsection{Influence of Different Time Steps}
Here, we evaluate the impact of different time steps on the performance of CP-DSA. The network structure used in the experiments is ResNet-14, with the initial time steps set to 2 and an interval of 2 between experimental time steps. From the results in Fig. \ref{fig4}, it can be observed that our model achieves good performance even with shorter time steps, thanks to the change-perceptive mechanism.
\subsection{Validation of the Effectiveness of the Change Perceptive Mechanism}
To verify the effectiveness of the proposed change perception mechanism used in the DSA model, we conduct ablation experiments on CP-DSA. For CIFAR10-DVS and DVS-Gesture, We select the ResNet-14 network architecture and perform experiments at time step 5. For N-Caltech101, we select ResNet-20 and set the time step to 8. The results are shown in Table \ref{tab6}. It can be observed that the CP mechanism has a positive effect on all three datasets, and using the EH and WK mechanisms alone also provides certain improvements. It can be observed that using the change perception mechanism on the DSA model resulted in improvements of 4.8\%, 3.28\%, and 3.48\% on three datasets, respectively. The effects of using the enhancement perception or weaken perception mechanisms alone were inferior to those of using the change perception mechanism.
\begin{figure*}[t]
\centering
\includegraphics[width=0.8\textwidth]{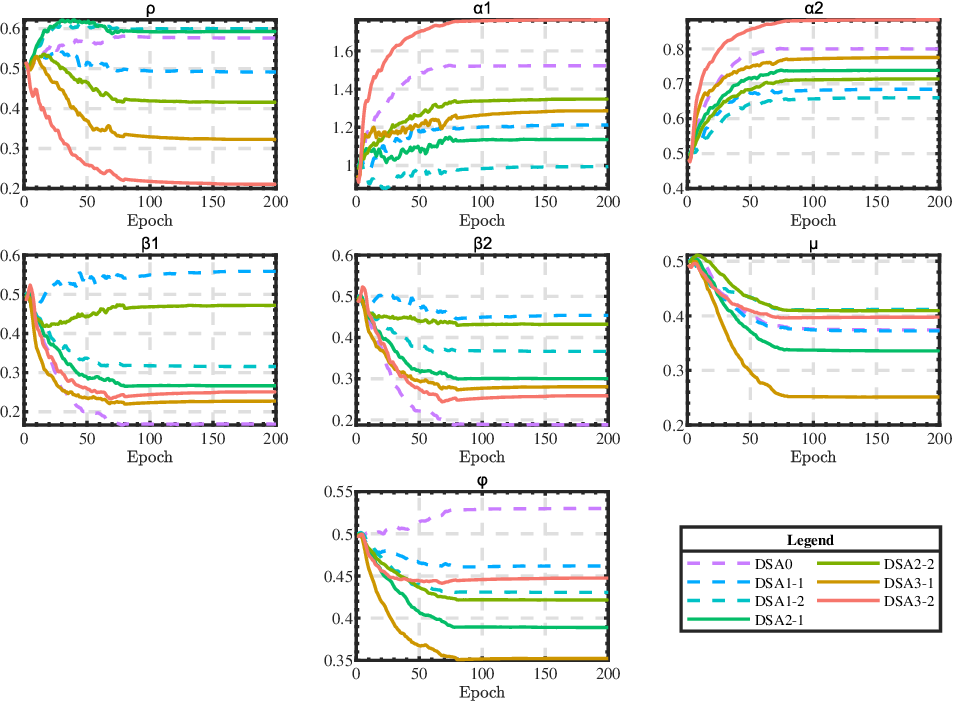} 
\caption{The hyperparameter variation tracking curve of the CP-DSA neurons. Labels are used to distinguish neurons in different layers, which are represented by curves of different colors in the graph. DSA0 corresponds to the neurons in the direct encoding layer, while DSA1-1 represents the first neuron in the sequence of the first convolutional structure. These hyperparameters gradually converge after 100 generations.}
\label{fig5}
\end{figure*}
\begin{table}[t]
\centering
\caption{The impact of three different perception mechanisms on the DSA model.}
\resizebox{0.48\textwidth}{!}{
\begin{tabular}{c c c c}
\toprule
Method & CIFAR10-DVS & N-Caltech101 & DVS-Gesture\\
\midrule
DSA & 77.0 & 79.22 & 89.58\\
+EH & 77.8$_{+0.8}$ & 79.47$_{+0.25}$ & 90.63$_{+1.05}$\\
+WK & 80.2$_{+3.2}$ & 81.23$_{+1.98}$ & 92.36$_{+2.78}$\\
+CP & \textbf{81.8}$_{+4.8}$ & \textbf{82.5}$_{+3.28}$ & \textbf{93.06}$_{+3.48}$\\
\bottomrule
\end{tabular}}
\label{tab6}
\end{table}

\subsection{Hyperparameter Tracking}
In this section, experiments are conducted using the ResNet-14 network on the DVS-Gesture dataset. Fig. \ref{fig5} illustrates the variation curves of the trainable parameters for CP-DSA neurons at different layers throughout the iterative process. In the figure, DSA0 corresponds to the neurons in the direct encoding layer, DSA1-1 corresponds to the neurons used in the first convolution layer of the first convolutional structure, and so on. From the curve variations, it can be observed that all hyperparameters ultimately stabilize and converge to a constant value, demonstrating the stability of CP-DSA. Additionally, the same parameters at different layers tend to converge to different values, confirming the adaptive capability of CP-DSA neurons. This enhances the representational power of spiking neural networks, thereby improving the network's performance.

\section{Discussion}
This paper, inspired by the structure of biological neuronal connections, introduces simplified dendritic temporal dynamics to the LIF neuron, resulting in the DSA neuron with powerful temporal learning capabilities. However, the introduction of new variables inevitably increases the computational burden.

More and more researchers are focusing on the attention mechanisms in Spiking Neural Networks (SNNs). However, the attention inherently carried by the neuron itself has not been fully explored. Inspired by conditioned reflexes, we focus on the information conveyed by changes in the neuronal membrane potential. By modeling this into the somatic potential of the neuron, we observe unexpected results: through the adaptive computation of the potential difference between adjacent time steps, the neuron can achieve excellent performance with short time steps (low latency) without the need for additional complex network modules or the introduction of extra neuronal dynamics mechanisms. The quantitative form of the change-perception mechanism may appear simple, but it is highly effective.

Additionally, inspired by the MLF method \cite{feng2022multi}, it naturally follows that the change-perception mechanism couples with the DSA neuron, leading to impressive performance. A concern, however, is whether the multi-spike level structure might lead to increased energy consumption.

Finally, we hope to inspire researchers to continue exploring the role of adjacent difference information in Spiking Neural Networks. Optimizing low-latency Spiking Neural Networks is necessary. In the future, we hope that with advances in biology, researchers will refine dendritic and somatic dynamics modeling to enhance the performance of spiking neurons. Most importantly, we look forward to testing the real-world performance of the CP-DSA neuron on neuromorphic hardware, evaluating its computational time and energy consumption to fully verify the superiority of CP-DSA.

\section{Conclusion}
In this work, we propose a novel neuron model called the Change-Perceptive Dendrite-Soma-Axon (CP-DSA) Neuron, which is designed for neuromorphic data (or temporal data). The CP-DSA consists of two parts. The first part is the DSA neuron, which is inspired by the biological neuronal connectivity structure. By introducing dendritic potentials and their temporal memory capabilities, the neuron gains a larger representational space for learning. The DSA neuron adopts an adaptive soft-reset mechanism, which corrects the information loss problem associated with hard resets while maintaining adaptability. The second part is the CP mechanism, which is inspired by classical conditioning. It regulates the neuron’s spike firing by calculating the changes between adjacent time steps and quantifying them into the soma potential. CP-DSA employs a parallel structure similar to MLF\cite{feng2022multi}, fully leveraging the CP mechanism while alleviating the gradient mismatch issue. Extensive experiments demonstrate the effectiveness of CP-DSA. We hope that CP-DSA will inspire the design of more high-performance and biologically interpretable spiking neural networks.

\bibliographystyle{IEEEtran}
\bibliography{IEEEabrv,ref}

\vfill
\end{document}